\documentclass[10pt,twocolumn,letterpaper]{article}

\usepackage{iccv}
\usepackage{times}
\usepackage{epsfig}
\usepackage{graphicx}
\usepackage{amsmath}
\usepackage{amssymb}

\usepackage{multirow}
\usepackage{verbatim}
\usepackage{color}
\usepackage{enumitem}

\usepackage[breaklinks=true,bookmarks=false]{hyperref}

\iccvfinalcopy 

\def\iccvPaperID{2862} 

\ificcvfinal\fi

\begin{document}

\title{Crowd Counting with Deep Structured Scale Integration Network}

\author{Lingbo Liu$^1$, Zhilin Qiu$^1$, Guanbin Li$^1$, Shufan Liu$^3$, Wanli Ouyang$^{3}$ and Liang Lin$^{1,2}$\thanks{Corresponding author is Liang Lin. This work was supported in part by the National Key Research and Development Program of China under Grant No.2018YFC0830103, in part by the National Natural Science Foundation of China under Grant No.U1811463 and No.61976250, in part by National High Level Talents Special Support Plan (Ten Thousand Talents Program). This work was also sponsored by SenseTime Research Fund.}\\
$^1$Sun Yat-sen University, China  {~~~} $^2$DarkMatter AI Research \\$^3$The University of Sydney, SenseTime Computer Vision Research Group, Australia
}

\maketitle
\ificcvfinal\thispagestyle{empty}\fi

\begin{abstract}
Automatic estimation of the number of people in unconstrained crowded scenes is a challenging task and one major difficulty stems from the huge scale variation of people.
In this paper, we propose a novel Deep Structured Scale Integration Network (DSSINet) for crowd counting, which addresses the scale variation of people by using structured feature representation learning and hierarchically structured loss function optimization.
Unlike conventional methods which directly fuse multiple features with weighted average or concatenation, we first introduce a Structured Feature Enhancement Module based on conditional random fields (CRFs) to refine multiscale features mutually with a message passing mechanism.
In this module, each scale-specific feature is considered as a continuous random variable and passes complementary information to refine the features at other scales.
Second, we utilize a Dilated Multiscale Structural Similarity loss to enforce our DSSINet to learn the local correlation of people's scales within regions of various size, thus yielding high-quality density maps.
Extensive experiments on four challenging benchmarks well demonstrate the effectiveness of our method. Specifically, our DSSINet achieves improvements of 9.5\% error reduction on Shanghaitech dataset and 24.9\% on UCF-QNRF dataset against the state-of-the-art methods.
\end{abstract}

\section{Introduction}

\begin{figure}[t]
\centering
   \includegraphics[width=0.85\columnwidth]{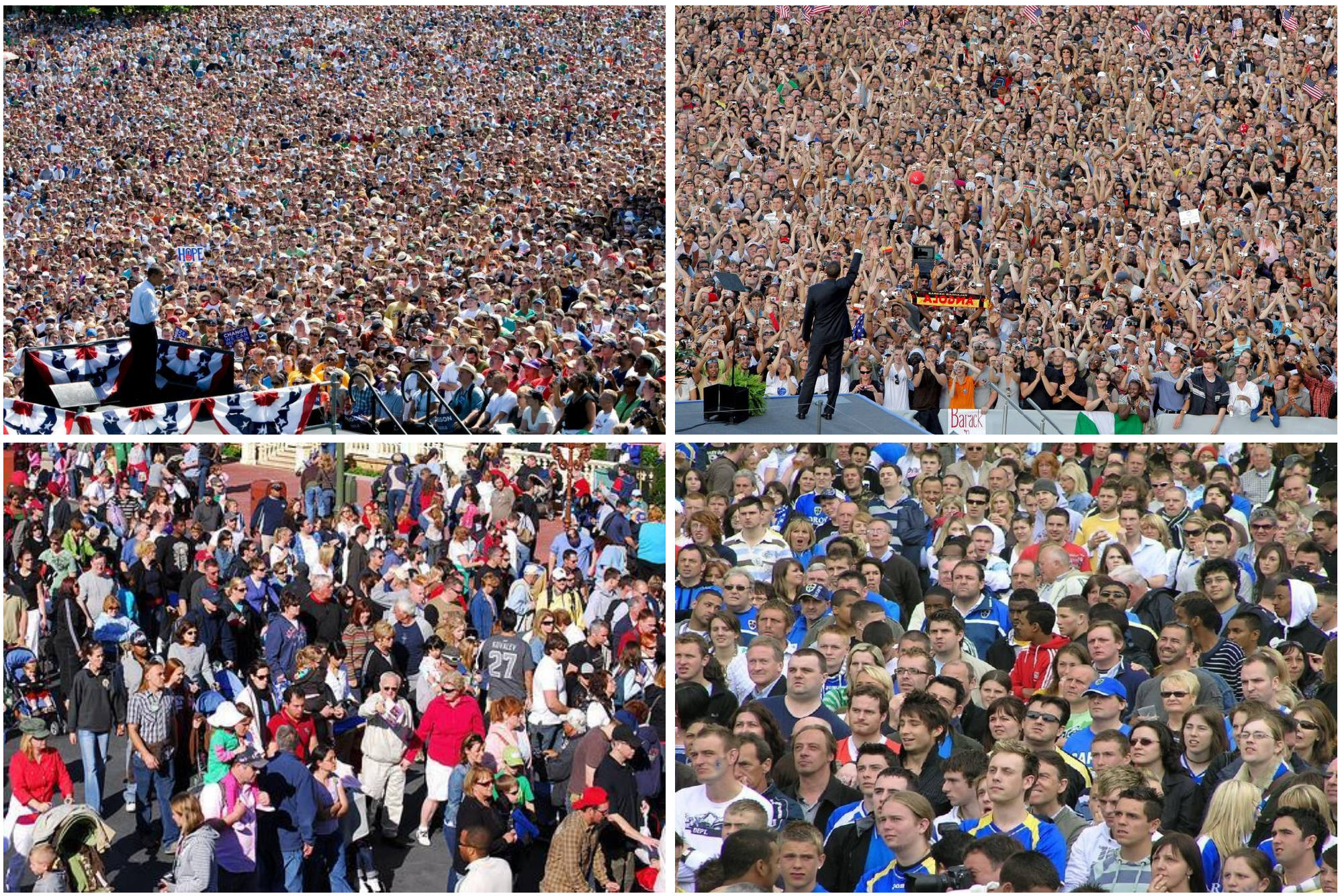}
\vspace{-1mm}
   \caption{Visualization of people with various scales in unconstrained crowded scenes. The huge scale variation of people is a major challenge limiting the performance of crowd counting.
   }
\vspace{-4mm}
\label{fig:challenge}
\end{figure}

Crowd counting, which aims to automatically generate crowd density maps and estimate the number of people in unconstrained scenes, is a crucial research topic in computer vision. Recently, it has received increasing interests in both academic and industrial communities, due to its wide range of practical applications, such as video surveillance \cite{xiong2017spatiotemporal}, traffic management \cite{zhang2017understanding} and traffic forecast \cite{liu2018attentive}.

Although numerous models~\cite{ge2009marked,idrees2013multi} have been proposed, crowd counting remains a very challenging problem and one major difficulty originates from the huge scale variation of people. As shown in Fig.~\ref{fig:challenge}, the scales of people vary greatly in different crowded scenes and capturing such huge scale variation is non-trivial.
Recently, deep neural networks have been widely used in crowd counting and have made
substantial progress~\cite{zhang2015cross,onoro2016towards,zeng2017multi,liu2018context,zhang2018crowd,liu2018adcrowdnet,qiu2019crowd}.
To address scale variation, most previous works utilized multi-column CNN~\cite{zhang2016single} or stacked multi-branch blocks~\cite{cao2018scale} to extract multiple features with different receptive fields, and then fused them with weighted average or concatenation.
However, the scale variation of people in diverse scenarios is still far from being fully solved.

In order to address scale variation, our motivations are two-fold.
{\bf{First}}, features of different scales contain different information and are highly complementary. For instance, the features at deeper layers encode high-level semantic information, while the features at shallower layers contain more low-level appearance details. Some related researches~\cite{xu2017learning,zhang2018bi} have shown that these complementary features can be mutually refined and thereby become more robust to scale variation.
However, most existing methods use simple strategies (e.g., weighted average and concatenation) to fuse multiple features, and can not well capture the complementary information. Therefore, it is very necessary to propose a more effective mechanism for the task of crowd counting, to fully exploit the complementarity between different scale features and improve their robustness.

\begin{figure}[t]
\centering
   \includegraphics[width=0.875\columnwidth]{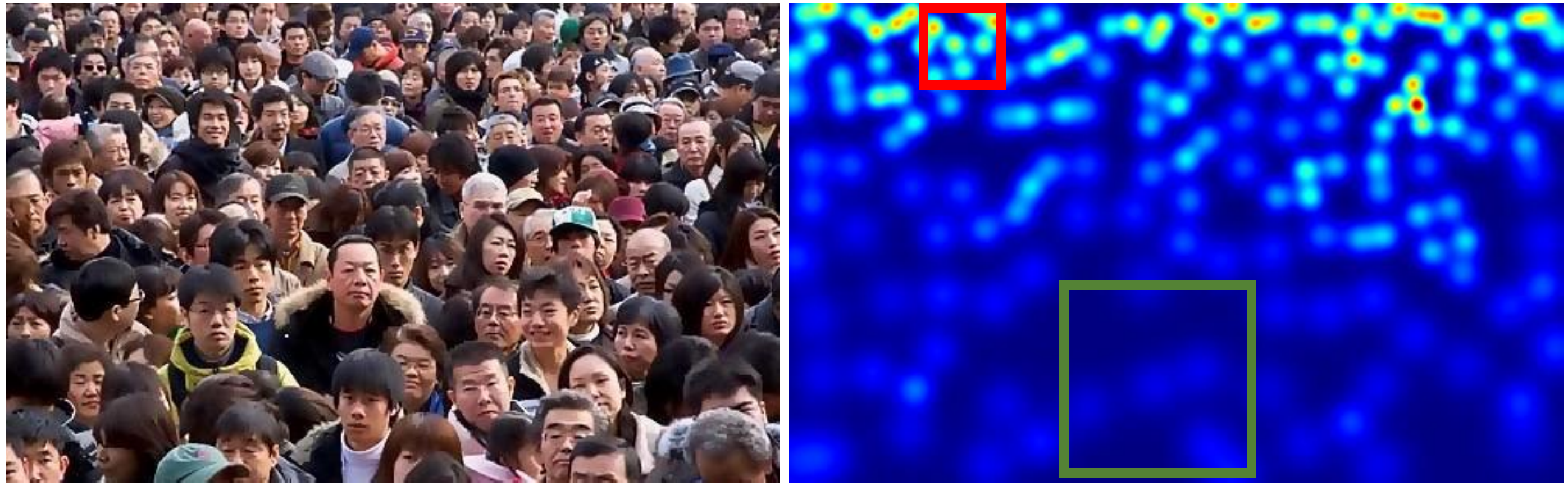}
\vspace{-1.0mm}
   \caption{Illustration of the information of people's scales on crowd density maps. The radiuses of people's heads are relatively uniform in a local region. Moreover, the local correlation of people's scales may vary in different regions.
   }
\vspace{-3mm}
\label{fig:scale_information}
\end{figure}

{\bf{Second}}, crowd density maps contain rich information of the people's scales\footnote{In this paper, ground-truth crowd density maps are generated with geometry-adaptive Gaussian kernels~\cite{zhang2016single}. Each person is marked as a Gaussian kernel with individual radius}, which can be captured by an effective loss function.
As shown in Fig.~\ref{fig:scale_information}, people in a local region usually have similar scales and the radiuses of their heads are relatively uniform on density maps, which refers to the local correlation of people's scales in our paper.
Moreover, this pattern may vary in different locations.
{\bf{i)}} In the area near the camera, the radiuses of people's heads are large and their density values are consistently low, thus it's better to capture the local correlation of this case in a large region (See the green box in Fig.~\ref{fig:scale_information}).
{\bf{ii)}} In the place far away from the camera, the heads' radiuses are relatively small and the density map is sharp, thus we could capture the local correlation of this case in a small region (See the red box in Fig.~\ref{fig:scale_information}).
However, the commonly used pixel-wise Euclidean loss fails to adapt to these diverse patterns.
Therefore, it is desirable to design a structured loss function to model the local correlation within regions of different sizes.

In this paper, we propose a novel Deep Structured Scale Integration Network (DSSINet) for high-quality crowd density maps generation, which addresses the scale variation of people from two aspects, including structured feature representation learning and hierarchically structured loss function optimization.
First, our DSSINet consists of three parallel subnetworks with shared parameters, and each of them takes a different scaled version of the same image as input for feature extraction.
Then, a unified Structured Feature Enhancement Module (SFEM) is proposed and integrated into our network for structured feature representation learning.
Based on the \textit{conditional random fields} (CRFs~\cite{lafferty2001conditional}), SFEM mutually refines the multiscale features from different subnetworks with a message passing mechanism~\cite{krahenbuhl2011efficient}.
Specifically, SFEM dynamically passes the complementary information from the features at other scales to enhance the scale-specific feature.
Finally, we generate multiple side output density maps from the refined features and obtain a high-resolution density map in a top-down manner.

For the hierarchically structured loss optimization, we utilize a Dilated Multiscale Structural Similarity (DMS-SSIM) loss to enforce networks to learn the local correlation within regions of various sizes and produce locally consistent density maps.
Specifically, our DMS-SSIM loss is designed for each pixel and it is computed by measuring the structural similarity between the multiscale regions centered at the given pixel on an estimated density map and the corresponding regions on ground-truth.
Moreover, we implement the DMS-SSIM loss with a dilated convolutional neural network, in which the dilated operation enlarges the diversity of the scales of local regions and can further improve the performance.

In summary, the contributions of our work are three-fold:
\begin{itemize}
\vspace{-0mm}
\setlength{\itemsep}{0pt}
\setlength{\parsep}{0pt}
\setlength{\parskip}{1pt}
  \item We propose a CRFs-based Structured Feature Enhancement Module, which refines multiscale features mutually and boosts their robustness against scale variation by fully exploiting their complementarity.
  \item We utilize a Dilated Multiscale Structural Similarity loss to learn the local correlation within regions of various sizes. To our best knowledge, we are the first to incorporate the MS-SSIM~\cite{wang2003multiscale} based loss function for crowd counting and verify its effectiveness in this task.
 \item Extensive experiments conducted on four challenging benchmarks demonstrate that our method achieves state-of-the-art performance.
\end{itemize}

\begin{figure*}[t]
\begin{center}
 \includegraphics[width=1.975\columnwidth]{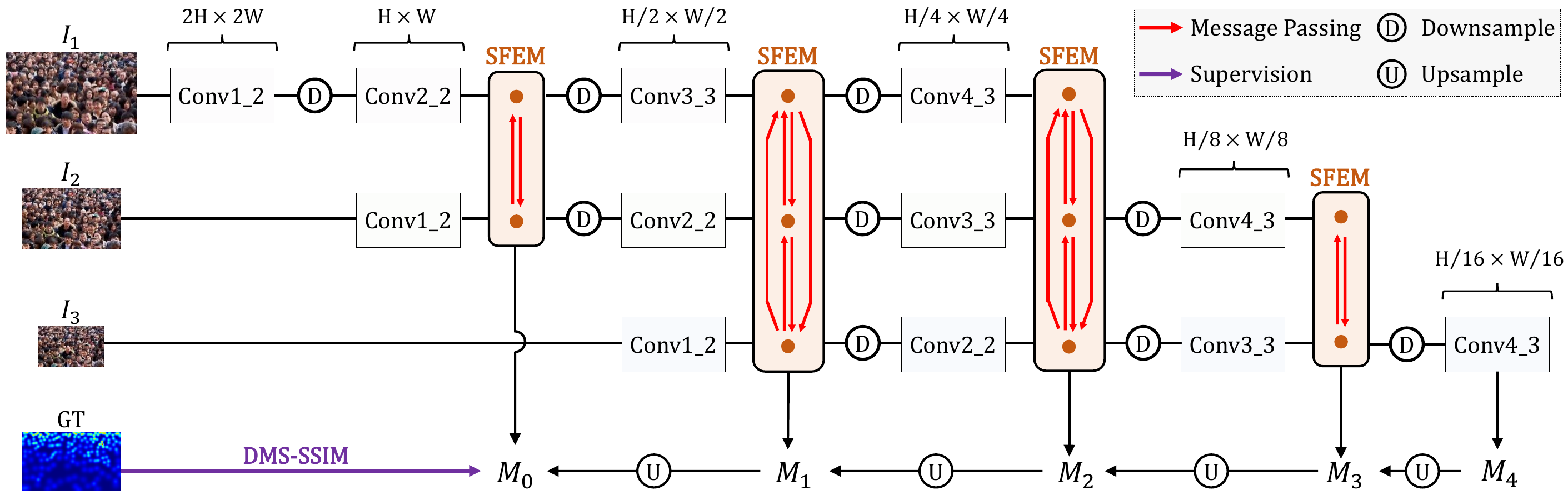}
 \vspace{-4mm}
\end{center}
   \caption{The overall framework of the proposed Deep Structured Scale Integration Network (DSSINet).
   DSSINet consists of three parallel subnetworks with shared parameters. These subnetworks take different scaled versions of the same image as input for feature extraction.
   First, DSSINet integrates four CRFs-based Structured Feature Enhancement Modules ({\bf{SFEM}}) to refine the multiscale features from different subnetworks.
   Then, we progressively fuse multiple side output density maps to obtain a high-resolution one in a top-down manner.
   Finally, a Dilated Multiscale Structural Similarity ({\bf{DMS-SSIM}}) loss is utilized to optimize our DSSINet.
   }
\vspace{-2mm}
\label{fig:model}
\end{figure*}

\section{Related Work}
{\bf{Crowd Counting:}}
Numerous deep learning based methods~\cite{walach2016learning,sam2017switching,xiong2017spatiotemporal,liu2018decidenet,liu2018crowd,liu2019geometric,gao2019pcc} have been proposed for crowd counting.
These methods have various network structures and the mainstream is a multiscale architecture, which extracts multiple features from different columns/branches of networks to handle the scale variation of people.
For instance, Boominathan et al.~\cite{boominathan2016crowdnet} combined a deep network and a shallow network to learn scale-robust features. Zhang et al.~\cite{zhang2016single} developed a multi-column CNN to generate density maps. HydraCNN~\cite{onoro2016towards} fed a pyramid of image patches into networks to estimate the count. CP-CNN~\cite{sindagi2017generating} proposed a Contextual Pyramid CNN to incorporate the global and local contextual information for crowd counting. Cao et al.~\cite{cao2018scale} built an encoder-decoder network with multiple scale aggregation modules.
However, the issue of the huge variation of people's scales is still far from being fully solved.
\textit{In this paper, we further strengthen the robustness of DSSINet against the scale variation of people from two aspects, including structured feature representation learning and hierarchically structured loss function optimization.}

{\bf{Conditional Random Fields:}}
In the field of computer vision, CRFs have been exploited to refine the features and outputs of convolutional neural networks (CNN) with a message passing mechanism~\cite{krahenbuhl2011efficient}. 
For instance, Zhang et al.~\cite{zheng2015conditional} used CRFs to refine the semantic segmentation maps of CNN by modeling the relationship among pixels. Xu et al.~\cite{xu2017learning} fused multiple features with Attention-Gated CRFs to produce richer representations for contour prediction.
Wang et al.~\cite{wang2018dividing} introduced an inter-view message passing module based on CRFs to enhance the view-specific features for action recognition. \textit{In crowd counting, we are the first to utilize CRFs to mutually refine multiple features at different scales and prove its effectiveness for this task.}

{\bf{Multiscale Structural Similarity:}}
MS-SSIM~\cite{wang2003multiscale} is a widely used metric for image quality assessment. Its formula is based on the luminance, contrast and structure comparisons between the multiscale regions of two images.
In \cite{zhao2017loss}, MS-SSIM loss has been successfully applied in image restoration tasks (e.g., image denoising and super-resolution), but its effectiveness has not been verified in high-level tasks (e.g, crowd counting).
Recently, Cao et al.~\cite{cao2018scale} combined Euclidean loss and SSIM loss~\cite{wang2004image} to optimize their network for crowd counting, but they can only capture the local correlation in regions with a fixed size.
\textit{In this paper, to learn the local correlation within regions of various sizes, we modify MS-SSIM loss with a dilated operation and show its effectiveness in this high-level task.}

\section{Method}
In this section, we propose a Deep Structured Scale Integration Network (DSSINet) for crowd counting.
Specifically, it addresses the scale variation of people with structured feature representation learning and structured loss function optimization.
For the former, a Structured Feature Enhancement Module based on \textit{conditional random fields} (CRFs) is proposed to refine multiscale features mutually with a message passing mechanism.
For the latter, a Dilated Multiscale Structural Similarity loss is utilized to enforce networks to learn the local correlation within regions of various sizes.

\subsection{DSSINet Overview}\label{sec:overview}
In crowd counting, multiscale features are usually extracted to handle the scale variation of people. Inspired by \cite{lin2017feature,kang2018crowd,chen2018deeplab}, we build our DSSINet with three parallel subnetworks, which have the same architecture and share parameters.
As shown in Fig.~\ref{fig:model}, these subnetworks are composed of the first ten convolutional layers (up to Conv4\_3) of VGG16~\cite{simonyan2014very} and each of them takes a different scaled version of the same image as input to extract features. Unlike previous works~\cite{zhang2016single,cao2018scale} that simply fuse features by weighted averaging or concatenation, our DSSINet adequately refines the multiscale features from different subnetworks.

Given an image of size ${H\times W}$, we first build a three-levels image pyramid $\{I_1, I_2, I_3\}$, where $I_2 \in R^{H\times W}$ is the original image, and $I_1 \in R^{2H\times 2W}$ and $I_3 \in R^{\frac {H}{2}\times \frac {W}{2}}$ are the scaled ones.
Each of these images is fed into one of the subnetworks. The feature of image $I_k$ at the Conv$i\_j$ layer of VGG16 is denoted as $f^k_{i,j}$. We then group the features with the same resolution from different subnetworks and form four sets of multiscale features $\{f^1_{2,2}, f^2_{1,2}\}$, $\{f^1_{3,3}, f^2_{2,2}, f^3_{1,2}\}$, $\{f^1_{4,3}, f^2_{3,3}, f^3_{2,2}\}$, $\{f^2_{4,3}, f^3_{3,3}\}$.
In each set, different features complement each other, because they are inferred from different receptive fields and are derived from different convolutional layers of various image sizes.
For instance, $f^2_{1,2}$ mainly contains the appearance details and $f^1_{2,2}$ encodes some high-level semantic information.
To improve the robustness of scale variation, we refine the features in the aforementioned four sets with the Structured Feature Enhancement Module described in Section~\ref{sec:sfem}. With richer information, the enhanced feature $\hat{f}^k_{i,j}$ of $f^k_{i,j}$ becomes more robust to the scale variation.
Then, $\hat{f}^k_{i,j}$ is fed into the following layer of $k^{th}$ subnetwork for deeper feature representation learning.

After structured feature learning, we generate a high-resolution density map in a top-down manner.
First, we apply a $1\times1$ convolutional layer on top of the last feature $f^3_{4,3}$ for reducing its channel number to 128, and then feed the compressed feature into a $3\times3$ convolutional layer to generate a density map $M_4$. However, with a low resolution of $\frac {H}{16}\times \frac {W}{16}$, $M_4$ lacks spatial information of people.
To address this issue, we generate other four side output density maps $\tilde{M}_0, \tilde{M}_1, \tilde{M}_2, \tilde{M}_3$ at shallower layers, where $\tilde{M}_i$ has a resolution of $\frac {H}{2^i}\times \frac {W}{2^i}$.
Specifically, $\tilde{M}_3$ is computed by feeding the concatenation of $f^2_{4,3}$ and $f^3_{3,3}$ into two stacked convolutional layers. The first $1\times1$ convolutional layer is also utilized to reduce the channel number of the concatenated feature to 128, while the second $3\times3$ convolutional layer is used to regress $\tilde{M}_3$. $\tilde{M}_2, \tilde{M}_1, \tilde{M}_0$ are obtained in the same manner.
Finally, we progressively pass the density maps at deeper layers to refine the density maps at shallower layers and the whole process can be expressed as:
\begin{small}
\setlength\abovedisplayskip{3.5pt}
\setlength\belowdisplayskip{3.5pt}
\begin{equation}
M_i = w_{i}*\tilde{M}_{i} + w_{i+1} * Up(M_{i+1}), i=3,2,1,0
\end{equation}
\end{small}%
where $w_{i}$ and $w_{i+1}$ are the parameters of two $3\times3$ convolutional layers and $Up()$ denotes a bilinear interpolation operation with a upsampling rate of 2. $M_i$ is the refined density map of $\tilde{M}_{i}$. The final crowd density map $M_0 \in R^{H\times W}$ has fine details of the spatial distribution of people.

Finally, we train our DSSINet with the Dilated Multiscale Structural Similarity (DMS-SSIM) loss described in Section~\ref{sec:msssim}.
We implement DMS-SSIM loss with a lightweight dilated convolutional neural network with fixed Gaussian-kernel and the gradient can be back-propagated to optimize our DSSINet.

\subsection{Structured Feature Enhancement Module}\label{sec:sfem}

In this subsection, we propose a unified Structured Feature Enhancement Module (SFEM) to improve the robustness of our feature for scale variation.
Inspired by the dense prediction works~\cite{chu2016crf,xu2017learning}, our SFEM mutually refines the features at different scales by fully exploring their complementarity with a \textit{conditional random fields} (CRFs) model.
In this module, each scale-specific feature passes its own information to features at other scales. Meanwhile, each feature is refined by dynamically fusing the complementary information received from other features.

Let us denote multiple features extracted from different subnetworks as $F=\{f_1, f_2,...,f_n\}$. $F$ can be any of the multiscale features sets defined in Section~\ref{sec:overview}.
Our objective is to estimate a group of refined features $\hat{F}=\{\hat{f}_1, \hat{f}_2,...,\hat{f}_n\}$, where $\hat{f}_i$ is the corresponding refined feature of $f_i$. We formulate this problem with a CRFs model. Specifically, the conditional distribution of the original feature $F$ and the refined feature $\hat{F}$ is defined as:
{\begin{small}
\setlength\abovedisplayskip{3pt}
\setlength\belowdisplayskip{3pt}
\begin{equation}
P(\hat{F} | F,\Theta) = \frac{1}{Z(F)} exp\{E(\hat{F},F,\Theta)\},
\end{equation}
\end{small}}%
where $Z(F)=\int_{\hat{F}} exp\{E(\hat{F},F,\Theta)\} d\hat{F}$ is the partition function for normalization and $\Theta$ is the set of parameters.
The energy function $E(\hat{F},F,\Theta)$ in CRFs is defined as:
{\begin{small}
\setlength\abovedisplayskip{3pt}
\setlength\belowdisplayskip{3pt}
\begin{equation}
E(\hat{F},F,\Theta) = \sum_i \phi(\hat{f}_i, f_i) + \sum_{i,j}\psi(\hat{f}_i, \hat{f}_j). \\
\end{equation}
\end{small}}%
In particular, the unary potential $\phi(\hat{f}_i, f_i)$, indicating the similarity between the original feature and the refined feature, is defined as:
{\begin{small}
\setlength\abovedisplayskip{3pt}
\setlength\belowdisplayskip{3pt}
\begin{equation}
\phi(\hat{f}_i, f_i) = -\frac{1}{2} ||\hat{f}_i - f_i||^2.
\end{equation}
\end{small}}%
We model the correlation between two refined features with a bilinear potential function, thus the pairwise potential is defined as:
{\begin{small}
\setlength\abovedisplayskip{3pt}
\setlength\belowdisplayskip{3pt}
\begin{equation}
\psi(\hat{f}_i, \hat{f}_j) = (\hat{f}_i)^T{}w^i_j{~}\hat{f}_j,
\end{equation}
\end{small}}%
where $w^i_j$ is a learned parameter used to compute the relationship between $\hat{f}_i$ and $\hat{f}_j$.

This is a typical formulation of CRF and we solve it with mean-field inference~\cite{ristovski2013continuous}. The feature $\hat{f}_i$ is computed by:
{\begin{small}
\setlength\abovedisplayskip{3pt}
\setlength\belowdisplayskip{3pt}
\begin{equation}
\hat{f}_i = f_i +\sum_{j\neq i}w^i_j{~}\hat{f}_j,
\label{crf}
\end{equation}
\end{small}}%
where the unary term is feature $f_i$ itself and the second term denotes the information received from other features at different scales. The parameter $w^i_j$ determines the information content passed from $f_j$ to $f_i$.
As $\hat{f}_i$ and $\hat{f}_j$ are interdependent in Eq.(\ref{crf}), we obtain each refined feature iteratively with the following formulation:
{\begin{small}
\setlength\abovedisplayskip{3pt}
\setlength\belowdisplayskip{3pt}
\begin{equation}
h_i^0 = f_i, {~~~~} h_i^t = f_i +\sum_{j\neq i}w^i_j{~}h_j^{t-1},  t=1{~}to{~}n,{~~~~} \hat{f}_i = h_i^n,\\
\label{crf_iteration}
\end{equation}
\end{small}}%
where $n$ is the total iteration number and $h_i^t$ is the intermediate feature at $t^{th}$ iteration.
The Eq.(\ref{crf_iteration}) can be easily implemented in our SFEM. Specifically, we apply a $1\times1$ convolutional layer to pass the complementary information from $f_j$ to $f_i$. $w^i_j$ is the learned parameter of the convolutional layer and it is shared for all iterations.

As shown in Fig.~\ref{fig:model}, we apply the proposed SFEM to mutually refine the features in $\{f^1_{2,2}, f^2_{1,2}\}$, $\{f^1_{3,3}, f^2_{2,2}, f^3_{1,2}\}$, $\{f^1_{4,3}, f^2_{3,3}, f^3_{2,2}\}$, $\{f^2_{4,3}, f^3_{3,3}\}$. After receiving the information from other features at different scales, the refined feature becomes more robust to the scale variation of people. The experiments in Section~\ref{sec:experiment} show that our SFEM greatly improves the performance of crowd counting.

\subsection{Dilated Multiscale Structural Similarity Loss}\label{sec:msssim}
In this subsection, we employ a Dilated Multiscale Structural Similarity (DMS-SSIM) loss to train our network.
We ameliorate the original MS-SSIM~\cite{wang2003multiscale} with dilation operations to enlarge the diversity of the sizes of local regions and force our network to capture the local correlation within regions of different sizes.
Specifically, for each pixel, our DMS-SSIM loss is computed by measuring the structural similarity between the multiscale regions centered at the given pixel on an estimated density map and the corresponding regions on the GT density map.

\begin{figure}
\begin{center}
\includegraphics[width=0.45\textwidth]{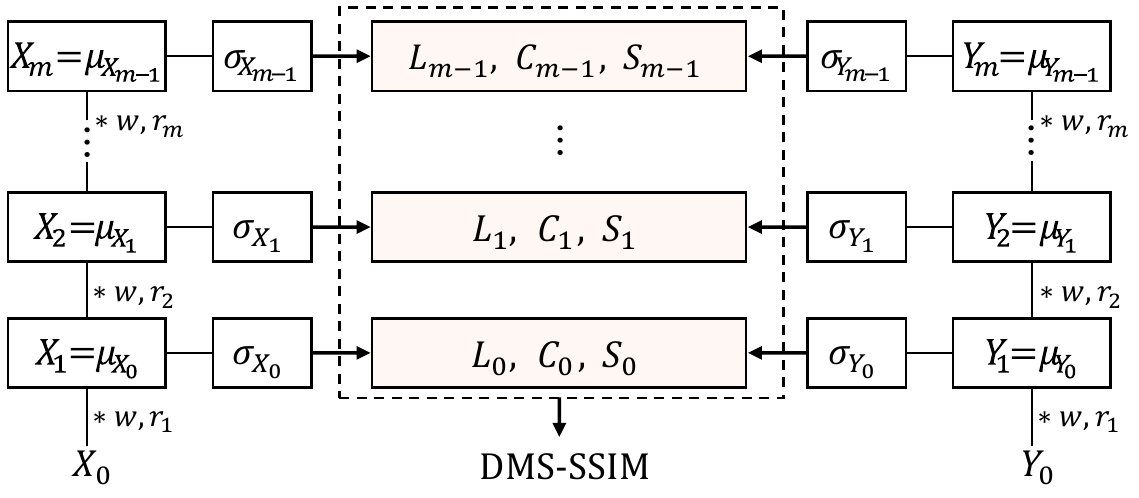}
\vspace{-1.75mm}
\caption{The network of Dilated Multiscale Structural Similarity loss. The normalized Gaussian kernel $w$ is fixed and shared for all layers. $r_i$ is the dilation rate of the $i^{th}$ layer.
$X_0$ and $Y_0$ are the estimated density map and the corresponding GT map respectively.
}
\label{fig:MSSSIM}
\vspace{-5mm}
\end{center}
\end{figure}

As shown in Fig.~\ref{fig:MSSSIM}, we implement the DMS-SSIM loss with a dilated convolutional neural network named as DMS-SSIM network. The DMS-SSIM network consists of $m$ dilated convolutional layers and the parameters of these layers are fixed as a normalized Gaussian kernel with a size of $5\times5$ and a standard deviation of 1.0. The Gaussian kernel is denoted as $w = \{w(o) | o\in O, O=\{(-2,-2),...,(2,2)\}\}$, where $o$ is an offset from the center.

For convenience, the estimated density map $M_0$ in DSSINet is re-marked as $X_0$ in this subsection and its corresponding ground-truth is denoted as $Y_0$.
We feed $X_0$ and $Y_0$ into the DMS-SSIM network respectively and their outputs at $i^{th}$ layer are represented as $X_i \in R^{H \times W}$ and $Y_i \in R^{H \times W}$.
Specifically, for a given location $p$, $X_{i+1}(p)$ is calculated by:
{\begin{small}
\setlength\abovedisplayskip{3pt}
\setlength\belowdisplayskip{3pt}
\begin{equation}
X_{i+1}(p) = \sum_{o\in O}w(o) \cdot  X_{i}(p+ r_{i+1} \cdot o),
\label{local_mean}
\end{equation}
\end{small}}%
where $r_{i+1}$ is the dilation rate of the $(i+1)^{th}$ layer and it is used to control the size of receptive field. Since $\sum_{o\in O}w(o)$ = 1, $X_{i+1}(p)$ is the weighted mean of a local region centered at the location $p$ on $X_{i}$ and we could get $X_{i+1}=\mu_{X_{i}}$.
In this case, $X_1$ is the local mean of $X_0$ and $X_2$ is the local mean of $X_1$. By analogy, $X_{i+1}(p)$ can be considered as the mean of a relatively large region on $X_0$.
Based on the filtered map $X_{i+1}$ , we calculate the local variance $\sigma^2_{X_{i}}(p)$ as:
{
\begin{small}
\setlength\abovedisplayskip{3pt}
\setlength\belowdisplayskip{3pt}
\begin{equation}
\sigma^2_{X_{i}}(p) = \sum_{o\in O} w(o)\cdot [{X_{i}}(p+r_{i+1} \cdot o) - \mu_{X_{i}}(p)]^2.
\label{local_variance}
\end{equation}
\end{small}%
}%
The local mean $\mu_{Y_{i}}$ and variance $\sigma^2_{Y_{i}}$ of filtered map $Y_{i}$ are also calculated with the same formulations as Eq.(\ref{local_mean}) and Eq.(\ref{local_variance}). Moreover, the local covariance $\sigma^2_{X_{i}Y_{i}}(p)$ between $X_{i}$ and $Y_{i}$ can be computed by:
{
\begin{small}
\setlength\abovedisplayskip{3pt}
\setlength\belowdisplayskip{3pt}
\begin{equation}
\begin{split}
\sigma^2_{X_{i}Y_{i}}(p) = \sum_{o\in O} \{w(o) &\cdot [X_{i}(p+r_{i+1} \cdot o) - \mu _{X_{i}}(p)] \\
&\cdot[Y_{i}(p+r_{i+1} \cdot o) - \mu_{Y_{i}}(p)]\}
\end{split}
\end{equation}
\end{small}
}

Further, the luminance comparison $L_i$, contrast comparison $C_i$ and structure comparison $S_i$ between $X_i$ and $Y_i$ are formulated as:
{\begin{footnotesize}
\setlength\abovedisplayskip{3pt}
\setlength\belowdisplayskip{3pt}
\begin{equation}
L_i={\frac {2\mu_{X_i}\mu_{Y_i}+c_{1}}{\mu_{X_i}^{2}+\mu_{Y_i}^{2}+c_{1}}},
C_i={\frac {2\sigma _{X_i}\sigma _{Y_i}+c_{2}}{\sigma _{X_i}^{2}+\sigma _{Y_i}^{2}+c_{2}}},
S_i={\frac {\sigma _{X_iY_i}+c_{3}}{\sigma _{X_i}\sigma _{Y_i}+c_{3}}},
\end{equation}
\end{footnotesize}}%
where $c_1$, $c_2$ and $c_3$ are small constants to avoid division by zero.
The SSIM between $X_i$ and $Y_i$ is calculated as:
{\begin{small}
\setlength\abovedisplayskip{3pt}
\setlength\belowdisplayskip{3pt}
\begin{equation}
\textit{SSIM}(X_i, Y_i) = L_i \cdot C_i \cdot S_i.
\end{equation}
\end{small}}%
Finally, the DMS-SSIM between the estimated density map $X_0$ and ground-truth $Y_0$, as well as the DMS-SSIM loss are defined as:
{
\begin{small}
\setlength\abovedisplayskip{3pt}
\setlength\belowdisplayskip{3pt}
\begin{equation}
\begin{split}
\textit{DMS-SSIM}(X_0, Y_0)&=
 \prod_{i=0}^{m-1} \{\textit{SSIM}(X_i, Y_i)\}^{\alpha_i}, \\
\textit{Loss}(X_0, Y_0)&=1-\textit{DMS-SSIM}(X_0, Y_0)
\label{ms-ssim_loss}
\end{split}
\end{equation}
\end{small}%
}%
where $\alpha_i$ is the importance weight of $\textit{SSIM}(X_i, Y_i)$ and we refer to \cite{wang2003multiscale} to set the value of $\alpha_i$.

In this study, the number of layers in DMS-SSIM network $m$ is set to 5. The dilation rates of these layers are 1, 2, 3, 6 and 9 respectively and we show the receptive field of each layer in Table~\ref{tab:receptive field}.
The receptive field of the $2^{nd}$ layer is 13, thus $\textit{SSIM}(X_1, Y_1)$ indicates the local similarity measured in regions with size of $13\times13$. Similarly, $\textit{SSIM}(X_4, Y_4)$ is the local similarity measured in $85\times85$ regions.
When removing the dilation, the receptive fields of all layers are clustered into a small region, which makes DMS-SSIM network fail to capture the local correlation in regions with large heads, such as the green box in Fig.~\ref{fig:scale_information}.
The experiments in Section~\ref{sec:ablation} show that the dilation operation is crucial to obtain the accurate count of crowd.

\begin{table}
\centering
\newcommand{\tabincell}[2]{\begin{tabular}{@{}#1@{}}#2\end{tabular}}
  \centering
  \resizebox{8.35cm}{!} {
    \begin{tabular}{c|c|c|c|c|c}
    \hline
    Layer & $1^{st}$ &$2^{nd}$ & $3^{rd}$ & $4^{th}$ & $5^{th}$\\
    \hline
    W/O Dilation   & 5 (1) & 9 (1)  & 13 (1) & 17 (1) & 21 (1) \\
    \hline
    W/- Dilation   & 5 (1) & 13 (2) & 25 (3) & 49 (6) & 85 (9) \\
    \hline
    \end{tabular}
  }
  \vspace{-1mm}
  \caption{Comparison of the receptive field in DMS-SSIM network with and without dilation. The receptive field $f_i$ and dilation rate $r_i$ of $i^{th}$ layer are denoted as ``$f_i{~}(r_i)$'' in the table. Without dilation ($r_i$ = 1 for all layers), the receptive fields of all layers are clustered into a small region, which makes DMS-SSIM network fail to capture the local correlation in regions with large heads.}
  \vspace{-1mm}
  \label{tab:receptive field}
\end{table}

\section{Experiments}\label{sec:experiment}

\subsection{Implementation Details}

{\bf{Ground-Truth Density Maps Generation:}}
In this work, we generate ground-truth density maps with the geometry-adaptive Gaussian kernels~\cite{zhang2016single}. For each head annotation $p_i$, we mark it as a Gaussian kernel $\mathcal{N}(p_i,{\sigma ^2})$ on the density map, where the spread parameter $\sigma$ is equal to 30\% of the mean distance to its three nearest neighbors. Moreover, the kernel is truncated within $3\sigma$ and normalized to an integral of $1$. Thus, the integral of the whole density map is equal to the crowd count in the image.

{\bf{Networks Optimization:}}
Our framework is implemented with PyTorch~\cite{paszke2017automatic}.
We use the first ten convolutional layers of the pre-trained VGG-16 to initialize the corresponding convolution layers in our framework. The rest of convolutional layers are initialized by a Gaussian distribution with zero mean and standard deviation of 1e-6.
At each training iteration, 16 image patches with a size of $224\times224$ are randomly cropped from images and fed into DSSINet.
We optimize our network with Adam~\cite{kingma2014adam} and a learning rate of 1e-5 by minimizing the DMS-SSIM loss.

\subsection{Evaluation Metric}
For crowd counting, {\itshape{Mean Absolute Error}} (MAE) and {\itshape{Mean Squared Error}} (MSE) are two metrics widely adopted to evaluate the performance. They are defined as follows:
{
\setlength\abovedisplayskip{3pt}
\setlength\belowdisplayskip{3pt}
\begin{equation}\resizebox{.875\hsize}{!}{$
   \text{MAE} = \frac{1}{N} \sum_{i=1}^{N} ||\hat{P}_i - P_{i}||,
   \text{MSE} = \sqrt{\frac{1}{N} \sum_{i=1}^{N} ||\hat{P}_i - P_{i}||^2} {~},
   $}
\end{equation}}
where $N$ is the total number of the testing images, $\hat{P}_i$ and $P_{i}$ are the estimated count and the ground truth count of the $i^{th}$ image respectively. Specifically, $\hat{P}_i$ is calculated by the integration of the estimated density map.

\subsection{Comparison with the State of the Art}
\begin{table}
\newcommand{\tabincell}[2]{\begin{tabular}{@{}#1@{}}#2\end{tabular}}
  \centering
    \begin{tabular}{c|c|c|c|c}
    \hline
    \multirow{2}{*}{Method} &
    \multicolumn{2}{c|}{Part A} &
    \multicolumn{2}{c}{Part B} \\
    \cline{2-5}
    & MAE  & MSE & MAE & MSE \\
    \hline\hline
    MCNN~\cite{zhang2016single} & 110.2 & 173.2 & 26.4 & 41.3\\
    \hline
    SwitchCNN~\cite{sam2017switching} & 90.4 & 135 & 21.6 & 33.4\\
     \hline
    CP-CNN~\cite{sindagi2017generating} & 73.6 & 106.4 & 20.1 & 30.1\\
    \hline
    DNCL~\cite{shi2018crowd} & 73.5 & 112.3 & 18.7 & 26.0\\
    \hline
    ACSCP~\cite{shen2018crowd} & 75.7 & 102.7 & 17.2 & 27.4 \\
    \hline
    IG-CNN~\cite{babu2018divide} & 72.5 & 118.2 & 13.6 & 21.1 \\
    \hline
    IC-CNN~\cite{ranjan2018iterative} & 68.5 & 116.2 & 10.7 & 16.0\\
    \hline
    CSRNet~\cite{li2018csrnet} & 68.2 & 115.0 & 10.6 & 16.0 \\
    \hline
    {{SANet}}~\cite{cao2018scale} & 67.0 & 104.5 & 8.4 & 13.6\\
    \hline
    Ours & {\bf\textcolor{red}{60.63}} & {\bf\textcolor{red}{96.04}} & {\bf\textcolor{red}{6.85}} & {\bf\textcolor{red}{10.34}}\\
    \hline
    \end{tabular}
  \vspace{0mm}
   \caption{Performance comparison on Shanghaitech dataset.
   }
  \vspace{-2mm}
  \label{tab:ShanghaiTech}
\end{table}

\begin{figure*}[t]
\begin{center}
 \includegraphics[width=1.950\columnwidth]{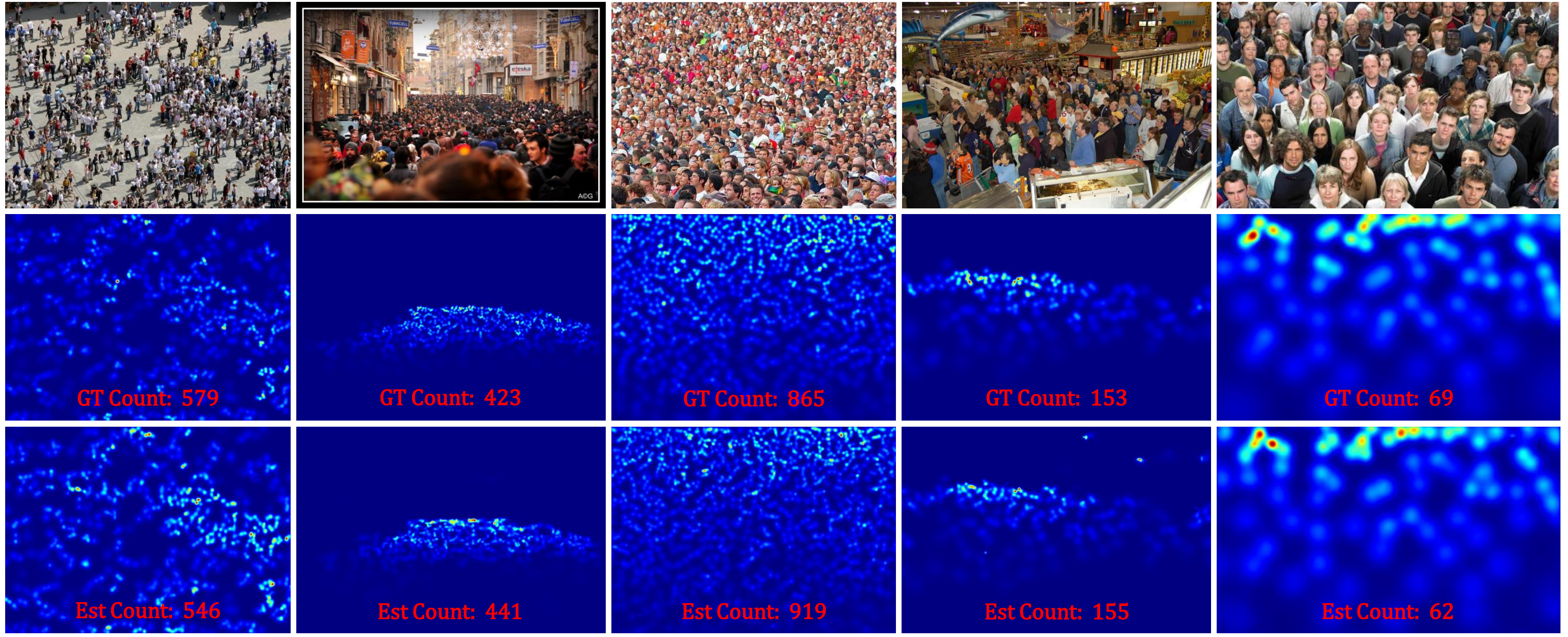}
 \vspace{-4mm}
\end{center}
   \caption{Visualization of the crowd density maps generated by our method on Shanghaitech Part A.
   The first row shows the testing images with people of various scales. The second row shows the ground-truth density maps and the standard counts, while the third row presents our generated density maps and estimated counts. Our method can generate high-quality crowd density maps with accurate counts, even though the scales of people vary greatly in images.
   }
\vspace{-1mm}
\label{fig:result_visual}
\end{figure*}

{\bf{Comparison on Shanghaitech~\cite{zhang2016single}: }}
As the most representative benchmark of crowd counting, Shanghaitech dataset contains 1,198 images with a total of 330 thousand annotated people. This dataset can be further divided into two parts: Part A with 482 images randomly collected from the Internet and Part B with 716 images taken from a busy shopping street in Shanghai, China.

We compare the proposed method with ten state-of-the-art methods on this dataset. As shown in Table~\ref{tab:ShanghaiTech}, our method achieves superior performance on both parts of the Shanghaitech dataset.
Specifically, on Part A, our method achieves a relative improvement of 9.5\% in MAE and 8.1\% in MSE over the existing best algorithm SANet~\cite{cao2018scale}.
Although previous methods have worked well on Part B, our method still achieves considerable performance gain by decreasing the MSE from 13.6 to 10.34.
The visualization results in Fig.~\ref{fig:result_visual} show that our method can generate high-quality crowd density maps with accurate counts, even though the scales of people vary greatly in images.

\begin{table}
\newcommand{\tabincell}[2]{\begin{tabular}{@{}#1@{}}#2\end{tabular}}
  \centering
    \begin{tabular}{c|c|c}
    \hline
    Method & MAE & MSE\\
    \hline
    \hline
    Idrees et al. ~\cite{idrees2013multi}   &  315 & 508 \\
    \hline
    MCNN~\cite{zhang2016single}                & 277 & 426 \\
    \hline
    Encoder-Decoder~\cite{badrinarayanan2015segnet} & 270 & 478 \\
    \hline
    CMTL~\cite{sindagi2017cnn}                    & 252 & 514\\
    \hline
    SwitchCNN~\cite{sam2017switching}                        & 228 & 445\\
    \hline
    Resnet-101~\cite{he2016deep}                        & 190 & 277\\
    \hline
    Densenet-201~\cite{huang2017densely}                        & 163 & 226\\
    \hline
    CL~\cite{idrees2018composition}                        & 132 & 191\\
    \hline
    Ours            & {\bf\textcolor{red}{99.1}} & {\bf\textcolor{red}{159.2}}\\
    \hline
    \end{tabular}
  \vspace{0mm}
    \caption{Performance of different methods on UCF-QNRF dataset.
    }
  \vspace{-0mm}
  \label{tab:UCF-QNRF}
\end{table}

{\bf{Comparison on UCF-QNRF~\cite{idrees2018composition}: }}
The recently released UCF-QNRF dataset is a challenging benchmark for dense crowd counting. It consists of 1,535 unconstrained crowded images (1,201 for training and 334 for testing) with huge scale, density and viewpoint variations. 1.25 million persons are annotated and they are unevenly dispersed to images, varying from 49 to 12,865 per image.

On UCF-QNRF dataset, we compare our DSSINet with eight methods, including Idrees et al.~\cite{idrees2013multi}, MCNN~\cite{zhang2016single}, Encoder-Decoder~\cite{badrinarayanan2015segnet}, CMTL~\cite{sindagi2017cnn}, SwitchCNN~\cite{sam2017switching}, Resnet-101~\cite{he2016deep}, Densenet-201~\cite{huang2017densely} and CL~\cite{idrees2018composition}.
The performances of all methods are summarized in Table~\ref{tab:UCF-QNRF} and we can observe that our DSSINet exhibits the lowest MAE 99.1 and MSE 159.2 on this dataset, outperforming other methods with a large margin.
Specifically, our method achieves a significant improvement of 24.9\% in MAE over the existing best-performing algorithm CL.

{\bf{Comparison on UCF\_CC\_50~\cite{idrees2013multi}: }}
This is an extremely challenging dataset. It contains 50 crowded images of various perspective distortions. Moreover, the number of people varies greatly, ranging from 94 to 4,543. Following the standard protocol in~\cite{idrees2013multi}, we divide the dataset into five parts randomly and perform five-fold cross-validation.
We compare our DSSINet with thirteen state-of-the-art methods on this dataset. As shown in Table~\ref{tab:UCF-CC}, our DSSINet obtains a MAE of 216.9 and outperforms all other methods. Specifically, our method achieves a relative improvement of 19.1\% in MAE over the existing best algorithm SANet~\cite{cao2018scale}.

\begin{table}
\newcommand{\tabincell}[2]{\begin{tabular}{@{}#1@{}}#2\end{tabular}}
  \centering
    \begin{tabular}{c|c|c}
    \hline
    Method & MAE & MSE\\
    \hline
    \hline
    MCNN~\cite{zhang2016single}                 & 377.6 & 509.1 \\ \hline
    SwitchCNN~\cite{sam2017switching}       & 318.1 & 439.2\\ \hline
    CP-CNN~\cite{sindagi2017generating}         & 295.8 & 320.9\\ \hline
    IG-CNN~\cite{babu2018divide}                & 291.4 & 349.4 \\ \hline
    ConvLSTM~\cite{xiong2017spatiotemporal}     & 284.5 & {\bf\textcolor{red}{297.1}} \\ \hline
    CSRNet~\cite{li2018csrnet}                  & 266.1 & 397.5\\ \hline
    IC-CNN~\cite{ranjan2018iterative}           & 260.9 & 365.5 \\ \hline
    SANet~\cite{cao2018scale}                   & {\bf\textcolor{blue}{258.4}} & 334.9 \\
    \hline
    Ours            & {\bf\textcolor{red}{216.9}} & {\bf\textcolor{blue}{302.4}}\\
    \hline
    \end{tabular}
  \vspace{0mm}
    \caption{Performance comparison on UCF\_CC\_50. The results of top two performance are highlighted in {\textcolor{red}{red}} and {\textcolor{blue}{blue}} respectively.
    }
  \vspace{-3mm}
  \label{tab:UCF-CC}
\end{table}

{\bf{Comparison on WorldExpo'10~\cite{zhang2015cross}:}}
As a large-scale crowd counting benchmark with the largest amount of images, WorldExpo'10 contains 1,132 video sequences captured by 108 surveillance cameras during the Shanghai WorldExpo 2010. Following the standard protocol in \cite{zhang2015cross}, we take 3,380 annotated frames from 103 scenes as training set and 600 frames from remaining five scenes as testing set. When testing, we only measure the crowd count under the given Region of Interest (RoI).

The mean absolute errors of our method and thirteen state-of-the-art methods are summarized in Table~\ref{tab:WorldExpo}. Our method exhibits the lowest MAE in three scenes and achieves the best performance with respect to the average MAE of five scenes.
Moreover, compared with those methods that rely on temporal information~\cite{xiong2017spatiotemporal} or perspective map~\cite{zhang2015cross}, our method is more flexible to generate density maps and estimate the crowd counts.

\begin{table}
\centering
\newcommand{\tabincell}[2]{\begin{tabular}{@{}#1@{}}#2\end{tabular}}
  \centering
  \resizebox{8.35cm}{!} {
    \begin{tabular}{c|c|c|c|c|c|c}
    \hline
    Method & S1 & S2 & S3 & S4 & S5 & Ave\\
    \hline\hline
    Zhang et al~\cite{zhang2015cross}          & 9.8	&   14.1 &  14.3 &  22.2 & 3.7	& 12.9\\
    \hline
    MCNN~\cite{zhang2016single}                & 3.4	&   20.6 &	12.9 &	13.0 & 8.1	& 11.6\\
    \hline
    ConvLSTM~\cite{xiong2017spatiotemporal}    & 7.1	&   15.2 &	15.2 &	13.9 & 3.5	& 10.9\\
    \hline
    SwitchCNN~\cite{sam2017switching}          & 4.4	&   15.7 &	10.0 &  11.0 & 5.9	& 9.4\\
    \hline
    DNCL~\cite{shi2018crowd}                   & 1.9    &   12.1 &  20.7 &   8.3 & 2.6  & 9.1 \\
    \hline
    CP-CNN~\cite{sindagi2017generating}        & 2.9	&   14.7 &	10.5 &	10.4 & 5.8	& 8.86\\
    \hline
    CSRNet~\cite{li2018csrnet}                 & 2.9    &   11.5 &   8.6 &  16.6 & 3.4  & 8.6 \\
    \hline
    SANet~\cite{cao2018scale}                  & 2.6    &   13.2 &   9.0 &  13.3 & 3.0  & 8.2 \\
    \hline
    DRSAN~\cite{liu2018crowd}                  & 2.6	&   11.8 &	10.3 &  10.4 & 3.7	& 7.76 \\
    \hline
    ACSCP~\cite{shen2018crowd}                 & 2.8    &  14.05 &   9.6 &   8.1 & 2.9  & 7.5 \\
    \hline
    Ours                           & {\bf{1.57}} & {\bf{9.51}} & {\bf{9.46}} & {\bf{10.35}} & {\bf{2.49}} & {\bf{6.67}}\\
    \hline
    \end{tabular}
  }
  \vspace{-1mm}
  \caption{MAE of different methods on the WorldExpo'10 dataset.}
  \vspace{-3mm}
  \label{tab:WorldExpo}
\end{table}

\subsection{Ablation Study}\label{sec:ablation}
{\bf{Effectiveness of Structured Feature Enhancement Module: }}
To validate the effectiveness of SFEM, we implement the following variants of our DSSINet:
{
\begin{itemize}[leftmargin=5mm]
\setlength{\itemsep}{0pt}
\setlength{\parsep}{0pt}
\setlength{\parskip}{0pt}

  \item {\bf{W/O FeatRefine:}} This model feeds the image pyramid $\{I_1,I_2,I_3\}$ into the three subnetworks, but it doesn't conduct feature refinement. It takes the original features to generate side output density maps. For example, $\tilde{M}_1$ is directly generated from $f^1_{3,3}$, $f^2_{2,2}$ and $f^3_{1,2}$.

  \item {\bf{ConcatConv FeatRefine:}} This model also takes $\{I_1,I_2,I_3\}$ as input and it attempts to refine multiscale features with concatenation and convolution. For instance, it feeds the concatenation of $f^1_{3,3},f^2_{2,2},f^3_{1,2}$ into a $1\times1$ convolutional layer to compute the feature $\hat{f}^1_{3,3}$.  $\hat{f}^2_{2,2}$ and $\hat{f}^3_{1,2}$ are obtained in the same manner.

  \item {\bf{CRF-$\mathbf{n}$ FeatRefine:}} This model uses the proposed CRFs-based SEFM to refine multiscale features from $\{I_1,I_2,I_3\}$. We explore the influence of the iteration number $\mathbf{n}$ in CRF, e.g., $\mathbf{n}$=1,2,3.

\end{itemize}
}

We train and evaluate all aforementioned variants on Shanghaitech Part A.
As shown in Table~\ref{tab:SFEM}, the variant ``W/O FeatRefine'' obtains the worst performance for the lack of features refinement.
Although ``ConcatConv FeatRefine'' can reduce the count error to some extent by simply refining multiple features, its performance is still barely satisfactory.
In contrast, our SFEM fully exploits the complementarity among multiscale features and mutually refines them with CRFs, and thus significantly boosts the performance. Specifically, our ``CRF-2 FeatRefine'' achieves the best performance. However, too longer iteration number of CRFs in SFEM would degrade the performance (See ``CRF-3 FeatRefine''), since those multiscale features may be excessively mixed and loss their own semantic meanings. Thus, the iteration number $n$ in CRFs is 2 in our final model.

\begin{table}
\newcommand{\tabincell}[2]{\begin{tabular}{@{}#1@{}}#2\end{tabular}}
  \centering
    \begin{tabular}{c|c|cc}
    \hline
    Method & MAE & MSE \\
    \hline\hline
    W/O FeatRefine           & 68.85 & 119.09\\
    \hline
    ConcatConv FeatRefine    & 67.11 & 110.87\\
    \hline
    CRF-1 FeatRefine         & 64.37 & 108.61\\
    \hline
    CRF-2 FeatRefine         & {\bf{60.63}} & {\bf{96.04}}\\
    \hline
    CRF-3 FeatRefine         & 63.80 & 103.05\\
    \hline
    \end{tabular}
  \vspace{0mm}
   \caption{Performance of different variants of DSSINet on Part A of Shanghaitech dataset.}
   \vspace{0mm}
  \label{tab:SFEM}
\end{table}

\begin{table}
\centering
\newcommand{\tabincell}[2]{\begin{tabular}{@{}#1@{}}#2\end{tabular}}
  \centering
    \begin{tabular}{c|c|c|c|c}
    \hline
    Scales & 1 & 1+0.5 & 2+1+0.5 & 2+1+0.5+0.25\\
    \hline
    MAE   & 70.94 & 64.67 & 60.63 & 61.13\\
    \hline
    \end{tabular}
  \caption{Ablation study of the scale number of image pyramid on Part A of Shanghaitech dataset.}
  \vspace{-2mm}
  \label{tab:scale}
\end{table}

{\bf{Influence of the Scale Number of Image Pyramid: }}
To validate the effectiveness of multiscale input, we train multiple variants of DSSINet with different scale number of image pyramid and summarize their performance in Table~\ref{tab:scale}.
Notice that the single-scale variant directly feeds the given original image into one subnetwork to extract features and generates the final density map from four side output density maps at Conv$1\_2$, Conv$2\_2$, Conv$3\_3$ and Conv$4\_3$. The term ``2+1+0.5'' denotes an image pyramid with scales 2, 1 and 0.5. Other terms can be understood by analogy.
We can observe that the performance gradually increases as the scale number increases and it is optimal with three scales. Since the computation was too large when the scale ratio was set to 4 or larger, we did not include more.

{\bf{Effectiveness of Dilated Multiscale Structural Similarity Loss:}}
In this subsection, we evaluate the effectiveness of the DMS-SSIM loss for crowd counting.
For the purpose of comparison, we train multiple variants of DSSINet with different loss functions, including Euclidean loss, SSIM loss and various configurations of DMS-SSIM loss. Note that a DMS-SSIM loss with $m$ scales is denoted as ``DMS-SSIM-$m$'' and its simplified version without dilation is denoted as ``MS-SSIM-$m$''.
The performances of all loss functions are summarized in Table~\ref{tab:Loss}.
We can observe that the performance of DMS-SSIM loss gradually improves, as the scale number $m$ increases. When adopting ``DMS-SSIM-5'', our DSSINet achieves the best MAE 60.63 and MSE 96.04, outperforming the models trained by Euclidean loss or SSIM loss.
We also implement a ``DMS-SSIM-6'' loss, in which the sixth layer has a dilation rate of 9 and it attempts to capture the local correlation in $121\times121$ regions. However, the people's scales may not uniform in such large regions, thus the performance of ``DMS-SSIM-6'' has slightly dropped, compared with ``DMS-SSIM-5''.
Moreover, the performance of MS-SSIM loss is worse than that of DMS-SSIM loss, since the receptive fields in MS-SSIM loss are intensively clustered into a small region, which makes our DSSINet fail to learn the local correlation of the people with various scales.
These experiments well demonstrate the effectiveness of DMS-SSIM loss.

\begin{table}
\newcommand{\tabincell}[2]{\begin{tabular}{@{}#1@{}}#2\end{tabular}}
  \centering
    \begin{tabular}{c|c|c}
    \hline
    Loss Function & MAE & MSE\\
    \hline
    \hline
    Euclidean   &  67.68 & 108.45\\
    \hline
    SSIM               & 74.60 & 133.64 \\
    \hline
    MS-SSIM-2                   & 73.21 & 125.05\\
    MS-SSIM-3                   & 67.46 & 114.79\\
    MS-SSIM-4                   & 64.80 & 109.26\\
    MS-SSIM-5                   & 63.51 & 103.81\\
    \hline
    DMS-SSIM-2                  & 73.33 & 121.87\\
    DMS-SSIM-3                  & 67.12 & 112.86\\
    DMS-SSIM-4                  & 62.90 & 105.14\\
    DMS-SSIM-5                  & {\bf{60.63}} & {\bf{96.04}}\\
    DMS-SSIM-6                  & 62.60 & 103.27\\
    \hline
    \end{tabular}
  \vspace{0mm}
    \caption{Performance evaluation of different loss functions on Part A of Shanghaitech dataset. ``DMS-SSIM-$m$'' denotes a DMS-SSIM loss with $m$ scales and ``MS-SSIM-$m$'' is the corresponding simplified version without dilation.
  }
  \vspace{-0mm}
  \label{tab:Loss}
\end{table}

{\bf{Complexity Analysis:}}
We also discuss the complexity of our method.
As the subnetworks in our framework have shared parameters and the kernel size of the convolutional layers in SFEM is $1\times1$, the proposed DSSINet only has 8.858 million parameters, 86.19\% (7.635 million) of which come from its backbone network (the first ten convolutional layers of VGG-16). As listed in Table~\ref{tab:complexity}, the number of parameters of our DSSINet is only half of that of the existing state-of-the-arts~(e.g.~CSRNet). Compared with these methods, our DSSINet achieves better performance with much fewer parameters.
During the testing phase, DSSINet takes 450 ms to process a 720$\times$576 frame from SD surveillance videos on an NVIDIA 1080 GPU. This runtime speed is already qualified for the needs of many practical applications, since people do not move so fast and not every frame needs to be analyzed.

\begin{table}
\centering
\newcommand{\tabincell}[2]{\begin{tabular}{@{}#1@{}}#2\end{tabular}}
  \centering
  \resizebox{8.35cm}{!} {
    \begin{tabular}{c|c|c|c|c}
    \hline
    Model & CP-CNN & SwitchCNN & CSRNet & Ours \\
    \hline
    Parameter & 68.4 & 15.11 & 16.26 & 8.85\\
    \hline
    \end{tabular}
  }
  \vspace{-1mm}
  \caption{Comparison of the number of parameters (in millions).}
  \vspace{-2mm}
  \label{tab:complexity}
\end{table}

\section{Conclusion}
In this paper, we develop a Deep Structured Scale Integration Network for crowd counting, which handles the huge variation of people's scales from two aspects, including structured feature representation learning and hierarchically structured loss function optimization.
First, a Structured Feature Enhancement Module based on \textit{conditional random fields} (CRFs) is proposed to mutually refine multiple features and boost their robustness.
Second, we utilize a Dilated Multiscale Structural Similarity Loss to force our network to learn the local correlation within regions of various sizes, thereby producing locally consistent estimation results.
Extensive experiments on four benchmarks show that our method achieves superior performance in comparison to the state-of-the-art methods.

{\small
\bibliographystyle{ieee_fullname}
\bibliography{egbib}
}

\end{document}